\documentclass[letterpaper, 10 pt, conference]{ieeeconf}

\IEEEoverridecommandlockouts                              

\usepackage{hyperref}
\usepackage{blindtext}
\usepackage{graphicx}
\usepackage{multirow}
\usepackage{xcolor}

\title{\LARGE \bf
Evaluating and Enhancing Trustworthiness of LLMs in Perception Tasks
}

\author{Malsha Ashani Mahawatta Dona$^1$, Beatriz Cabrero-Daniel$^1$, Yinan Yu$^2$, Christian Berger$^1$\\
$^1$\textit{University of Gothenburg} and $^2$\textit{Chalmers University of Technology}\\
Gothenburg, Sweden \\
\{malsha.mahawatta,beatriz.cabrero-daniel,christian.berger\}@gu.se, yinan@chalmers.se}

\begin{document}

\maketitle
\thispagestyle{empty}
\pagestyle{empty}

\begin{abstract}

Today's advanced driver assistance systems (ADAS), like adaptive cruise control or rear collision warning, are finding broader adoption across vehicle classes. Integrating such advanced, multimodal Large Language Models (LLMs) on board a vehicle, which are capable of processing text, images, audio, and other data types, may have the potential to greatly enhance passenger comfort. Yet, an LLM's hallucinations are still a major challenge to be addressed. In this paper, we systematically assessed potential hallucination detection strategies for such LLMs in the context of object detection in vision-based data on the example of pedestrian detection and localization. We evaluate three hallucination detection strategies applied to two state-of-the-art LLMs, the proprietary GPT-4V and the open LLaVA, on two datasets (Waymo/US and PREPER CITY/Sweden). Our results show that these LLMs can describe a traffic situation to an impressive level of detail but are still challenged for further analysis activities such as object localization. We evaluate and extend hallucination detection approaches when applying these LLMs to video sequences in the example of pedestrian detection. Our experiments show that, at the moment, the state-of-the-art proprietary LLM performs much better than the open LLM. Furthermore, consistency enhancement techniques based on voting, such as the Best-of-Three (BO3) method, do not effectively reduce hallucinations in LLMs that tend to exhibit high false negatives in detecting pedestrians. However, extending the hallucination detection by including information from the past helps to improve results.
\end{abstract}

\section{Introduction}

Advanced driver assistance systems (ADAS), increasingly common in passenger vehicles, aim at improving traffic safety. Such systems support the driver in traffic situations when pulling back from a parking lot supported by a rear collision warning (RCW), lane departure warning (LDW), or adaptive cruise control (ACC) for example. The operational design domains (ODDs) around such scenarios are typically well-defined, and the necessary sensors like cameras, radars, and ultrasound devices have proven to work reliably for decades.

Recent trends in the automotive industry such as SOAFEE~\cite{soaffee} now aim at significantly simplifying a vehicle's system architecture to contain powerful, centralized processing units with general-purpose multi-core CPUs and hardware accelerators like GPUs to prepare the vehicle for future applications.
These hardware accelerator-powered platforms can execute specialized neural networks (NN) necessary for ADAS and Autonomous Driving (AD) features, and are even capable to run Large Language Models (LLMs) locally without a powerful computation infrastructure backend in the cloud. 
Successful adoptions of LLMs are typically seen in knowledge-intensive tasks as well as traditional natural language processing (NLP) tasks~\cite{yang2023harnessing}. LLMs built into vehicles are envisioned to improve the in-vehicle human-machine interface (HMI)~\cite{hybridReasonongLLMinCars}, and BMW Group has demonstrated an in-car expert powered by LLMs, capable of engaging in human-like conversation with the passengers to respond to inquiries related to vehicle features~\cite{bmwWebsite, bmwCarExpert}. There are also commercial-level applications such as LINGO-2~\cite{LINGO2} that introduce a natural language for autonomous driving, supporting the concept of vehicles with built-in foundational models.

\subsection{Problem Domain and Motivation} 

Recently, such LLMs have been applied to computer vision tasks where so far, specialized NNs were trained and used (cf.~\cite{29_VLM_Ped_Detection,VLM_Definition}). When deploying such generic LLMs onto a powerful, centralized processing unit in vehicles, they can not only be used as a better HMI, but also to support a vehicle's perception stack (cf.~\cite{hybridReasonongLLMinCars}).

However, hallucinations generated by such LLMs are an important challenge inhibiting their safe application in automotive scenarios~\cite{halluciationsInAD}. Hallucinations, as defined by Huang et al.~are a phenomenon where models tend to generate nonsensical or unfaithful information compared to the provided context or real-world knowledge~\cite{huang2023survey_hallucinations}. LLMs may cause hallucinations for various reasons, such as the quality of training corpora, the complexity of the model, and the randomness of the answer generation strategies~\cite{reasons_hallucinations}.

Seeing LLMs potentially finding their way into vehicles, the question of ``How well can we trust an LLM within a vehicle?'' given the likelihood of misinformation generated by them~\cite{halluciationsInAD} becomes an urgent topic to address. Thus, the trustworthiness of LLMs must be ensured by detecting and mitigating hallucinations, especially for safety-critical systems such as ADAS/AD, which would impact the safety of passengers and pedestrians in general~\cite{safetyCriticalSystems}.

\subsection{Research Goal and Research Questions}
\label{sec:researchQuestions}

The research goal of our study is to systematically evaluate the performance of hallucination detection techniques when potentially considering the use of state-of-the-art LLMs as part of a perception stack to detect pedestrians in urban environments. We derive the following research questions:

\textbf{RQ-1:} What are potential hallucination detection strategies suggested in the literature?

\textbf{RQ-2:} How can hallucinations be characterized when applying LLMs to pedestrian detection and localization for ADAS/AD?

\textbf{RQ-3:} How can hallucination detection strategies be enhanced for use in ADAS/AD perception and monitoring systems?

\subsection{Contributions and Scope}

We provide an overview of strategies to detect hallucinations in the context of an object detection task in an ADAS/AD perception stack that is hypothetically supported by an LLM. 
Our main contribution is the assessment of hallucination detection strategies extracted from recent publications; we also include our own enhancements to these techniques. We apply two state-of-the-art LLMs, GPT-4V, and the open LLaVA, which we execute locally, to vision-based data of pedestrian situations from two recent datasets (Waymo/US~\cite{waymoDataset} and PREPER CITY/Sweden~\cite{revereDataset}). 
We consider hallucination detection as the first step to establish hallucination mitigation in an LLM-supported perception stack, laying the foundation for identifying and categorizing hallucinations to pave the way to enhanced and reliable models with better performance.

\subsection{Structure of the Paper}

The rest of the paper is organized as follows: Section~\ref{sec:relatedWork} reviews the related work related to hallucination detection, whereas Section~\ref{sec:methodology} explains the adopted methodology in detail. Results of the experiments are presented in Section~\ref{sec:results}, and the analysis and discussion of the findings are interpreted in Section~\ref{AnalysisAndDiscussion}. Section~\ref{sec:conclusion} concludes the paper.

\section{Related Work}
\label{sec:relatedWork}

Reducing and ideally eliminating hallucinations~\cite{hallucinations} that prevent misinformation is a major step in the process of ensuring the trustworthiness of an LLM-based system. Its importance elevates significantly when LLM-based systems are used within safety-critical systems. We review relevant related work for using LLMs in vision-oriented tasks relevant for automotive applications and potential hallucination detection strategies to address RQ-1.


The concept of using LLMs for visual recognition tasks, known as Vision-Language Models (VLMs)~\cite{VLM_Definition}, has emerged with the recent developments in the field of LLMs. Ranasinghe et al.~\cite{ranasinghe2024learning} elevate the Visual Question Answering (VQA) ability of the model by improving the spatial awareness in the VLM. Zhou et al.~\cite{VLM_zhou2023vision} explore the use of VLMs within the fields of Autonomous Driving (AD) and Intelligent Transportation Systems (ITS), focusing on the potential applications and research directions. These studies showcase that VLMs can be used within AD and ITS to integrate vision and language data to improve perception and understanding, for instance, in tasks such as pedestrian detection. Furthermore,~\cite{VLM_zhou2023vision} provide a detailed overview of LLMs and VLMs used in the field of AD with a taxonomy of tasks and types each model uses. This work can be considered as a comprehensive survey that explores the potential applications and emerging research directions in the context of VLMs.  

The studies from Liu et al.~\cite{29_VLM_Ped_Detection,30_vlm_PedDetection} propose different pedestrian detection strategies using VLM, showing that the usage of LLM within automotive is experimentally feasible. Also, the industry is reporting to explore the use of LLMs for potential commercialization~\cite{bmwCarExpert, bmwWebsite}. Research work such as ``Dolphins''~\cite{ma2023dolphins} accelerates the use of VLMs as driving assistants by proving how well a fully autonomous vehicle navigates through complex real-world scenarios using the assistance of a VLM. 


Hallucination detection and mitigation play a vital role in ensuring the trustworthiness of LLM-based systems. 
Manakul et al.~\cite{manakul2023selfcheckgpt} propose a hallucination detection strategy called ``SelfCheckGPT'' that utilizes a sampling approach to fact-check a model's responses. This method does not rely on any external databases or output probability distributions; in fact, it considers the simple logic that the answers must be consistent to some degree if the LLM has knowledge of a given concept. The proposed method was demonstrated to detect non-factual and factual sentences in the WikiBio dataset curated by the authors, but systematic evaluations in other application contexts are left open. A similar hallucination detection approach called DreamCatcher has been introduced by Liang et al.~\cite{Dreamcatcher} that uses consistency checking of multiple responses to detect hallucinations. In addition to these existing studies, there is also other research that used inconsistency checking to reduce the occurrence of hallucinations to some extent by initiating new chat completions with the LLMs without keeping history: for instance, Ronanki et al.~\cite{ronanki2022chatgpt} propose the ''Best of Three (BO3)'' strategy to improve the stability in the LLM generated responses in the domain of Agile Software Development, evaluating the quality of user stories. These proposed consistent checking mechanisms can be adapted into the domain of ADAS/AD to support perception tasks.


Galitsky introduces a fact-checking system called Truth-O-Meter to detect hallucinations in LLMs~\cite{truthOMetere}. This proposed method identifies incorrect facts by comparing the generated responses with the actual information scraped from other sources of information. Text mining and web scraping techniques are used to identify the corresponding correct sentences, whereas semantic generalization procedures are adapted to improve the generated content to mitigate the hallucinations. This method can be applied to vast genres of data, from sentences to phrases, including opinionated data. However, the proposed method can only be applied to the data to which correct and meaningful facts are available. 

In a more recent study, Wang et al.~\cite{HaELM} present``HaELM'', an LLM capable of detecting hallucinations thanks to its training with manually annotated hallucinated and non-hallucinated data obtained by Large Vision-Language Models such as ChatGPT. The model HaELM achieves lower accuracy and recall values than ChatGPT; however, it outperforms ChatGPT in precision and F1 Score. The concept behind HaELM is promising, but its applicability in a real-world setting would require training an LLM with manually annotated hallucinations in the selected domain. Furthermore, additional computational resources might be required for training purposes of the LLMs, which limits the applicability of this proposed detection strategy.

Many methods of hallucination detection have been proposed in previous studies. Yet, only a limited number of generalized evaluation benchmarks have been proposed to assess multimodal LLMs that cover text, images, and videos for example. Chen et al.~\cite{MHaluBench} have proposed the benchmark named MHaluBench and a novel hallucination detection framework named UNIHD to fill the aforementioned research gap. The proposed MHaluBench benchmark is capable of assessing different types of tasks covering different hallucination categories at a fine-grained level. The UNIHD framework detects hallucinations in the content generated by multimodal LLMs by gathering evidence from auxiliary tools. These auxiliary tools are selected by the queries crafted for core claims extracted from the initial responses. UNIHD is a task-agnostic framework that detects hallucinations, yet, an assessment for safety-critical application is left open.

Wang et al.~have also contributed to the field of hallucination evaluation by introducing AMBER~\cite{wang2024amber}, a multidimensional benchmark for hallucination evaluation. The proposed AMBER benchmark extracts the nouns of the initial response to generate an object list that will be checked against the annotations obtained by a different set of prompts. This benchmark requires a labelled dataset for discriminative tasks, which limits the applicability of the proposed methodology in hallucination evaluation in general. 

There is an emerging trend of research suggesting the systematic assessment of an LLM's output multiple times to ensure accuracy and reliability~\cite{manakul2023selfcheckgpt,Dreamcatcher}. However, these techniques often require complex prompting schemes, to which LLMs are sensitive~\cite{sensitivityPromptsLLM}. These prompts typically involve additional questions to probe the uncertainty or inconsistency of the answers. For our use case, we aim to use as simple prompts as possible to ensure robustness, which is crucial for automotive use cases. Hence, we have adopted the voting-based consistency enhancement strategy, BO3, and its adaptations by proposing a hallucination detection technique that is applicable in the ADAS/AD domain. We outline our approach as well as the systematic evaluation in Sec.~\ref{sec:methodology}.


\section{Methodology}
\label{sec:methodology}

\begin{figure*} [t!]
\centerline{\includegraphics[trim={2.6cm 1.5cm 3.5cm 1.65cm},clip,width=\linewidth]{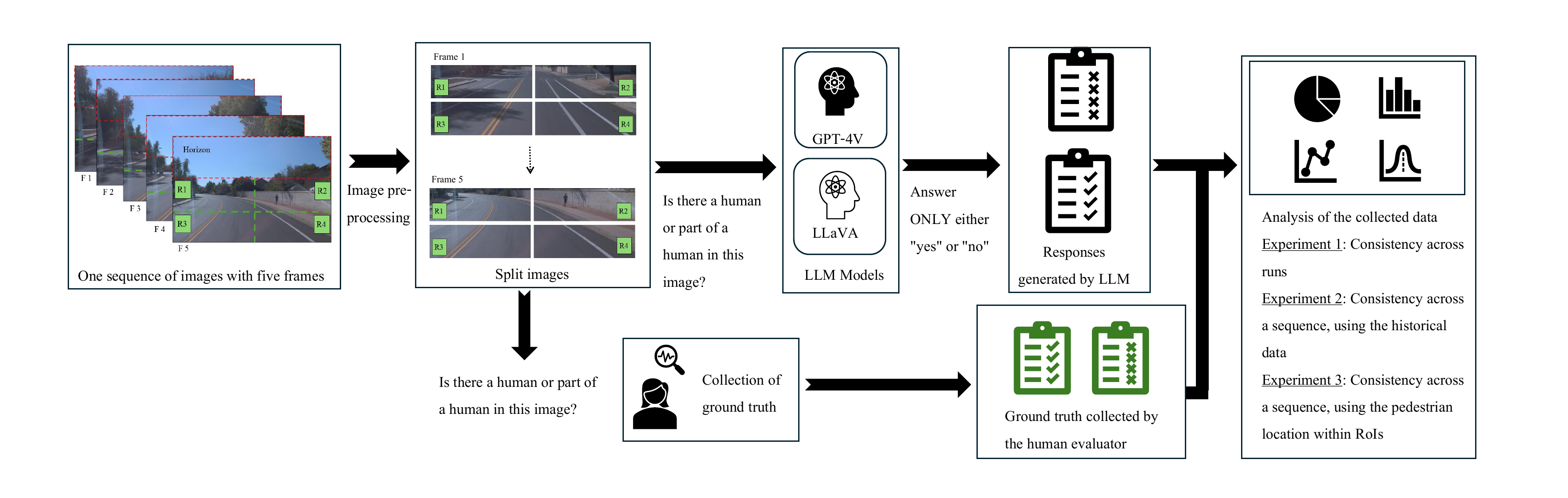} }%
\caption{Overview diagram of the experimental setup: All frames in each sequence of images are systematically cropped to remove the horizon and split into four RoIs to support the localization of pedestrians. All RoIs and full images are evaluated with the LLMs GPT-4V and LLaVA using the prompt: ``Is there a human or part of a human in this image? Answer ONLY either `yes' or `no'.'' The LLMs' responses are compared against the GT collected by human annotations to analyze the three experiments.}
\label{fig:overviewDiagram}
\end{figure*}


We selected image sequences of pedestrian scenarios that contain sequential data on the pedestrians and their trajectories over the duration, from two state-of-the-art datasets. For these sequences, we identified and labelled semantic Regions of Interest (RoIs) with their respective ground truth (GT). We then fed these images into two state-of-the-art LLMs with a predefined prompt and analyzed the results to answer our research questions. This section describes in detail our experiment pipeline, depicted in Fig.~\ref{fig:overviewDiagram}.

\subsection{Dataset Curation and Preparation}

We used images from different driving scenarios from the Waymo open dataset~\cite{waymoDataset} collected in urban and suburban areas in the USA, and from the PREPER CITY dataset collected by Yu et al.~\cite{revereDataset} in urban areas in Gothenburg, Sweden. We chose these two anonymized datasets to cover two different countries to reduce the impact of unknown effects originating from only using a single dataset. The Waymo motion dataset was extracted by sampling every ten frames from the front camera images in the training set, resulting in 15,947 images in our image pool. This dataset was collected in 2021, covering over 100,000 traffic segments. The PREPER CITY dataset was also collected in 2021, and the selected partition of the dataset contained 12,271 annotated images. 
As a first step, we curated the original datasets by selecting traffic scenarios that captured only a single pedestrian or a single cohesive group of pedestrians walking along or crossing the road. This constraint was imposed to reduce the impact of unknown effects originating from pedestrians scattered across an image, which could lead to an imbalanced dataset if every RoI contained a positive example. 
The final dataset used for experimentation contained 17 driving scenarios from Waymo and 18 driving scenarios from PREPER CITY. 
As the second step, the selected images were pre-processed to crop the horizon to ensure that the LLMs would only focus on the ground-level details. 
Assuming that the camera's optical axis is parallel to the road surface, the third step splits the images into four regions of interest (RoIs, labelled R1-R4 in Fig.~\ref{fig:overviewDiagram}) to identify the left, right, far, and close regions of the ground plane. This division is based on the assumption that the ego vehicle's orientation provides a consistent reference point, with left and right corresponding to the vehicle's lateral sides while far and close related to the distance from the camera lens.
We defined these RoIs to support the semantic localization of pedestrians as our preliminary experiments indicated that LLMs are not performing sufficiently reliably to localize pedestrians in a complete image but rather excel in describing an image's content. 
The final step of the data curation consisted of manually defining the GT of whether a (cohesive group of) pedestrian(s) was present. 
The resulting dataset contained 1868 RoIs in total, spread over 836 RoIs from Waymo and 1032 RoIs from PREPER CITY.


\subsection{Data Collection}

We used the curated dataset to feed our experiment pipeline, as shown in Fig.~\ref{fig:overviewDiagram}, to collect data about potential hallucinations. We conducted our experiments with the commercial gpt-4-vision-preview (GPT-4V) model and the locally installed Large Language-and-Vision Assistant (hallucinations) 1.6 model through the Ollama platform, and the results were compared across the models. For the locally executed LLM, we used the Ollama Python library to run the experiments with the LLaVA 1.6 model on an Intel Xeon W-1250 CPU with 128GB RAM and accelerated by an Nvidia Quadro P2200.

We fed the images from our curated dataset in randomized order before cropping and after cropping and splitting into the four RoIs to both LLMs along with the following prompt: \verb|Is there a human pedestrian in this image?| \verb|Answer only either "yes" or "no".| We presented the same image (or image's RoI) three times to an LLM to systematically assess the performance of the `Best of Three' (BO3) strategy across the spatial and temporal dimension, where the former refers to individual frames independently of each other, and the latter refers to consecutive frames to preserve a semantic context over time. We logged the Boolean replies from all experiments along with the complete execution time per prompt for our subsequent data analysis, resulting in 12,142 total responses. Furthermore, we created another set of boolean responses by labelling the images and RoIs randomly for all considered strategies. These randomly generated responses were used as the baseline when evaluating the performances of the two LLMs, GPT-4V and LLava. 


\subsection{Data Analysis}

We assessed the performance of the hallucination detection strategies using these responses as follows:

\textbf{DA-1:} We calculated performance metrics per complete image (ie., non-cropped) to evaluate the LLMs in detecting pedestrians and to explore potential hallucinations. This analysis contributes to answer RQ-2.

\textbf{DA-2:} We calculated performance metrics for all corresponding four RoIs of an image to assess the performance of the BO3 strategy. We also compared these results with the ones obtained from the unmodified images from DA-1 to evaluate whether cropping and splitting an image has an impact on an LLM's proneness for hallucinations. This analysis contributes to answering RQ-2 and RQ-3.

\textbf{DA-3:} We assessed the performance of the LLMs using sequences of consecutive frames in comparison to the individual frames to evaluate whether having temporal data on pedestrian visibility supports the detection of potential hallucinations. This analysis contributes to answer RQ-3.

\textbf{DA-4:} We evaluated to what extent splitting an image into smaller RoIs and semantically assessing an LLM's response considering pedestrian trajectory would help in detecting hallucinations. This analysis addresses RQ-3.

\section{Results} \label{sec:results}

In this section, we report the results from all experiments. For replication purposes, we share the necessary code and raw results together with explanatory video supplementary materials in a GitHub repository~\cite{Repogithub}.


\subsection{Types of Hallucinations and LLM Performance on Unmodified Images (DA-1, DA-2)}
\label{sec:TypesOfHallucinations+FullImagesSingleRUn}

\begin{figure} [t!]
\centerline{\includegraphics[width=\linewidth]{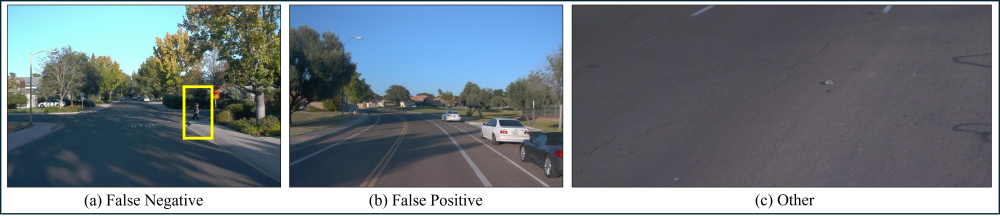} }%
\caption{Types of hallucinations: (a) False Negatives: The LLM is unable to detect the pedestrian in the left corner, highlighted by the yellow box. (b) False Positives: The LLM hallucinates a pedestrian in the image, whereas GT denotes that there is no human or a part of a human present in the image. (c) Other: The LLM refuses to process the picture, hallucinating some content that is not allowed by the safety system, resulting in a content policy violation.}
\label{fig:TypesofHallucinations}
\end{figure}

We have identified several types of hallucinations that are caused by LLMs (cf.~Figure~\ref{fig:TypesofHallucinations}) that are relevant in the ADAS/AD context, particularly for pedestrian detection. The two main types of hallucinations are false negatives and false positives, where false negatives are responses that report that there are no humans or parts of humans in the picture, whereas GT depicts a human or a part of a human (cf.~Figure~\ref{fig:TypesofHallucinations} part (a)). False positives are responses that contain hallucinations of humans or parts of humans in the images where there are no actual humans or parts of humans identified in the GT (cf.~Figure.~\ref{fig:TypesofHallucinations} part (b)). In addition to these, we saw another hallucination type where the model rejected processing the image due to a content policy violation where the model hallucinated some content that is not allowed to be processed by its safety system (cf.~Figure~\ref{fig:TypesofHallucinations} part (c)).       

Table~\ref{tab:fullimagesda1} shows the performance of the selected models GPT-4V and LLaVA, together with the responses that we randomly generated as response baseline. The recall and F1 scores are calculated for all responses and displayed for the two different datasets, Waymo and PREPER CITY, allowing to report about the sensitivity of the system and providing the balance between precision and recall.  

\begin{table}[]
\centering
\caption{Performance in terms of recall and F1-score for the two selected LLMs and for the randomly generated responses on the unmodified full images from the two selected datasets.}
\label{tab:fullimagesda1}
\begin{tabular}{l|l|l|l|l|}
\cline{2-5}
 & \multicolumn{2}{c|}{Waymo} & \multicolumn{2}{c|}{PREPER CITY} \\
 \cline{2-5}
& \multicolumn{1}{c|}{Recall} & \multicolumn{1}{c|}{F1 score} & \multicolumn{1}{c|}{Recall} & \multicolumn{1}{c|}{F1 score} \\ \hline
\multicolumn{1}{|l|}{GPT-4V} & 100.0\% & 90.13\% & 90.28\% & 89.97\% \\ \hline
\multicolumn{1}{|l|}{LLaVA} & 39.3\% & 51.97\% & 30.56\% & 41.31\% \\ \hline
\multicolumn{1}{|l|}{Random} & 48.26\% & 59.69\% & 50.0\% & 58.78\% \\ \hline
\end{tabular}
\end{table}

\subsection{BO3 Hallucination Detection across RoIs (DA-2)}

Table~\ref{tab:da2} displays the performance of GPT-4V, and LLaVA across the two datasets for the following four experiments:

\textbf{Single RoI}: Feeding a single frame to the LLM once and statistically analyzing the responses.

\textbf{BO3 per RoI}: Feeding a single frame three times to the LLM and statistically analyzing the responses.

\textbf{Two historical votes in sequences (THV)}: Feeding consecutive frames of the same RoI to the LLM and statistically analyzing the responses, preserving semantics over time.

\textbf{Two historical votes in sequences 2 (THV-2)}: Feeding consecutive frames of the same RoI to the LLM and analyzing the responses, checking the two sequential frames only when a pedestrian is not visible in the current frame but a pedestrian was detected in the two previous frames. 

We compared the performance of GPT-4 vision in the full images and in the four RoIs. It takes 8.1$\pm$5.1 seconds for GPT-4 vision to process a full image, while it takes 3.6$\pm$2.7 for GPT-4 vision to process each of the four RoIs. In turn, LLAVA took on average 3.7$\pm$1.0 seconds to process each of the four RoIs per frame. This tradeoff between the time spent per frame is countered by the benefit of locating the pedestrian in the image in a machine-interpretable way (as opposed to natural language descriptions).
The recall and F1 scores were calculated for each experiment and displayed under separate columns dedicated to each dataset in Table~\ref{tab:da2}. The ``Single RoI'' rows per each model show recall and F1 score value calculated based on the data collected by feeding single RoIs to the LLMs once. Similarly, the ``BO3 per RoI'' rows that merged into each model depict the statistics when the same RoI is tested three times against the LLM.

\begin{table}[]
\centering
\caption{Performance, in terms of recall and F1-score, of GPT-4V and LLaVA in labelling RoIs as containing a pedestrian or not. Three hallucination mitigation strategies (BO3, THV, and THV-2) are evaluated and compared to random labelling.}
\label{tab:da2}
\resizebox{\linewidth}{!}{
\begin{tabular}{ll|l|l|l|l|}
\cline{3-6}
 &  & \multicolumn{2}{c|}{Waymo} & \multicolumn{2}{c|}{PREPER CITY} \\ \cline{3-6}
 &  & \multicolumn{1}{c|}{Recall} & \multicolumn{1}{c|}{F1 score} & \multicolumn{1}{c|}{Recall} & \multicolumn{1}{c|}{F1 score} \\ \hline
\multicolumn{1}{|l|}{\multirow{4}{*}{GPT-4V}} & Single RoI & 98.62\% & 75.29\% &  92.47\% & 76.27\% \\ \cline{2-6} 
\multicolumn{1}{|l|}{} & BO3 per RoI & 98.34\% & 78.09\% &  93.55\% & 79.82\% \\ \cline{2-6} 
\multicolumn{1}{|l|}{} & THV & 87.87\% & 68.07\% &  79.46\% & 65.19\% \\ \cline{2-6} 
\multicolumn{1}{|l|}{} & THV-2 & 94.14\% & 64.19\% &  86.49\% & 62.99\% \\ \hline
\multicolumn{1}{|l|}{\multirow{4}{*}{LLaVA}} & Single RoI & 38.31\% & 29.75\% &  34.23\% & 27.76\% \\ \cline{2-6} 
\multicolumn{1}{|l|}{} & BO3 per RoI & 32.78\% & 27.82\% &  29.03\% & 26.21\% \\ \cline{2-6} 
\multicolumn{1}{|l|}{} & THV & 33.05\% & 26.87\% &  21.08\% & 18.8\% \\ \cline{2-6} 
\multicolumn{1}{|l|}{} & THV-2 & 49.79\% & 32.38\% & 38.92\% & 27.59\% \\ \hline
\multicolumn{1}{|l|}{\multirow{4}{*}{Random}} & Single RoI & 45.64\% & 29.27\% &  47.49\% & 29.53\% \\ \cline{2-6} 
\multicolumn{1}{|l|}{} & BO3 per RoI & 40.25\% & 26.01\% &  50.0\% & 31.26\% \\ \cline{2-6} 
\multicolumn{1}{|l|}{} & THV & 41.42\% & 25.98\% &  39.46\% & 25.75\% \\ \cline{2-6} 
\multicolumn{1}{|l|}{} & THV-2 & 59.41\% & 32.05\% & 58.92\% & 32.83\% \\ \hline
\end{tabular}
}
\end{table}

Table~\ref{tab:da2_consistency} contains the frequency of correct labels across the three runs for each RoI with the GT labels. For instance, when the GT is reporting no pedestrian in the image, the column header ``3 no'' would represent that the LLM correctly identifies that there is no pedestrian across all three tests. Similarly, ``2 no, 1 yes'' would represent that the LLM does not see the presence of the pedestrian two times but hallucinating a pedestrian once, whereas ``2 yes, 1 no'' would represent that the LLM is hallucinating a pedestrian twice but correctly not reporting the pedestrian once out of all three tests. The same representation has been used for the scenarios where the GT reports a pedestrian in the image. 

\begin{table}
\centering
\caption{Frequency of correct labels across test repetitions per RoI. }
\label{tab:da2_consistency}
\resizebox{\linewidth}{!}{
\begin{tabular}{|c|c|c|c|c|c|}
\hline
\textbf{LLM} & \textbf{GT label} & \textbf{3 no} & \textbf{2 no, 1 yes} & \textbf{2 yes, 1 no} & \textbf{3 yes} \\ \hline
GPT-4V & No & 1004 & 220 & 98 & 119 \\ \hline
LLaVA & No & 432 & 220 & 360 & 60 \\ \hline
\multicolumn{5}{c}{} \\ \hline
\textbf{LLM} & \textbf{GT label} & \textbf{3 yes} & \textbf{2 yes, 1 no} & \textbf{2 no, 1 yes} & \textbf{3 no} \\ \hline
GPT-4V & Yes & 401 & 10 & 6 & 10 \\ \hline
LLaVA & Yes & 27 & 106 & 175 & 119 \\ \hline
\end{tabular}
}
\end{table}

Further analysis of the collected data showed that LLMs are often inconsistent in answering the same question repeatedly: 17.89\% in the case of GPT-4V and 65.85\% in the case of LLaVA, ie., reporting varying answers for the same input.

\subsection{Hallucination Detection using Historical Frames in an Automotive Context (DA-3 and DA-4)}

When looking at the GT data, we see indications that sequential data, which contains information about previous frames, could help to detect hallucinations in the automotive context. This seems promising especially considering the false negatives when pedestrians were detected in previous frames. 

In the GT data, in 84.96\% of the frames where a pedestrian is seen, it is also seen in at least one of the previous two frames. This includes the transitions where an LLM correctly identifies the pedestrian in all two previous frames across the sequence, given that the pedestrian is visible in the third frame, and the transitions where an LLM identifies the pedestrian visibility correctly at least once in the two previous frames given the third frame has a visible pedestrian. 

On the other hand, only 5.37\% of the frames showing no pedestrian are preceded by two frames showing a pedestrian (e.g., the pedestrian disappears after two frames). Therefore, we can extract that this transition is an unlikely event to occur in real-world scenarios and use that understanding to check again if the LLM labels this.

\begin{figure}
    \centering
    \includegraphics[width=\linewidth]{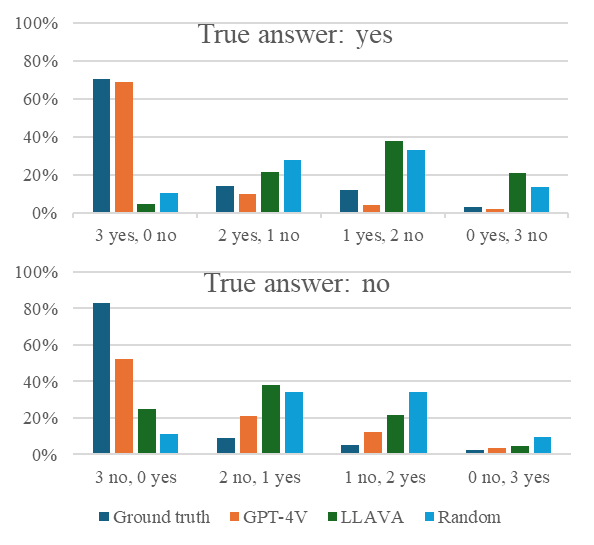}
    \vspace{-0.6cm}
    \caption{Proportion of yes/no labels among the RoI in focus and the same RoI in the previous two frames. The upper graph shows the percentage of `yes' labels in RoIs that show humans according to the GT, whereas the lower graph shows the percentage of `no' labels when that RoI does not contain a pedestrian. The navy bar, representing GT labels, shows that the labels for RoIs tend to remain the same across three consecutive frames.}
    \label{fig:bo3history2}
\end{figure}

Figure~\ref{fig:bo3history2} shows the distribution of each transition when sequential data are considered, focusing on three frames at a time. The two graphs hold the distribution percentages for GT, GPT-4V, LLaVA and Random, as described above.  



\textbf{Utilizing the Sequential Data for Hallucination Detection (DA-3):}
Due to the apparent importance of considering sequential data, two hallucination detection techniques were implemented across the image sequences focusing on RoIs to evaluate the performance of the LLMs for the selected two datasets as reported in Table~\ref{tab:da2} for THV and THV-2. 

The rows ``THV'' hold the recall and F1 score values for the evaluation of responses generated for the third frame, considering the responses of the previous two frames. If the majority vote is yes in the previous two frames, the third frame is identified as a frame with a pedestrian, and the response is evaluated against the GT. The same experiment was conducted for GPT-4V, LLaVA and Random to evaluate each model against Random, which is considered as a baseline distribution. The results show that 15.5\% of the frames that GPT4-V labelled as not showing a pedestrian were preceded by two frames labelled as showing a pedestrian (much higher probability than in the GT). This percentage increases to 27.51\% in the analysis by LLaVA. 

The rows ``THV-2'' represent the recall and F1 score values for the second approach that we adopted for experimentation in order to detect the hallucinations. In this approach, the majority vote was only considered to correct the label of the third frame, assigned by the LLM in the cases where it labels the image as no pedestrian. The results show that the THV-2 approach improves the overall performances of both the models GPT-4V (Recall to 94.14\%, 86.49\% and F1 score to 64.19\%, 62.99\%) and LLaVA (Recall to 49.79\%, 38.92\% and F1 score to 32.38\%, 27.59\%)for both datasets compared to the performances of the models under the THV approach. However, THV-2 and THV methods seem to not increase the performance of GPT-4V compared to LLaVA.

Similar to the THV-2 approach, alternative strategies could exploit historical data, for instance, by studying the probabilities of each of the transitions or sequences on a larger amount of data as described below.  

\begin{figure}
    \centering
    \includegraphics[width=\linewidth]{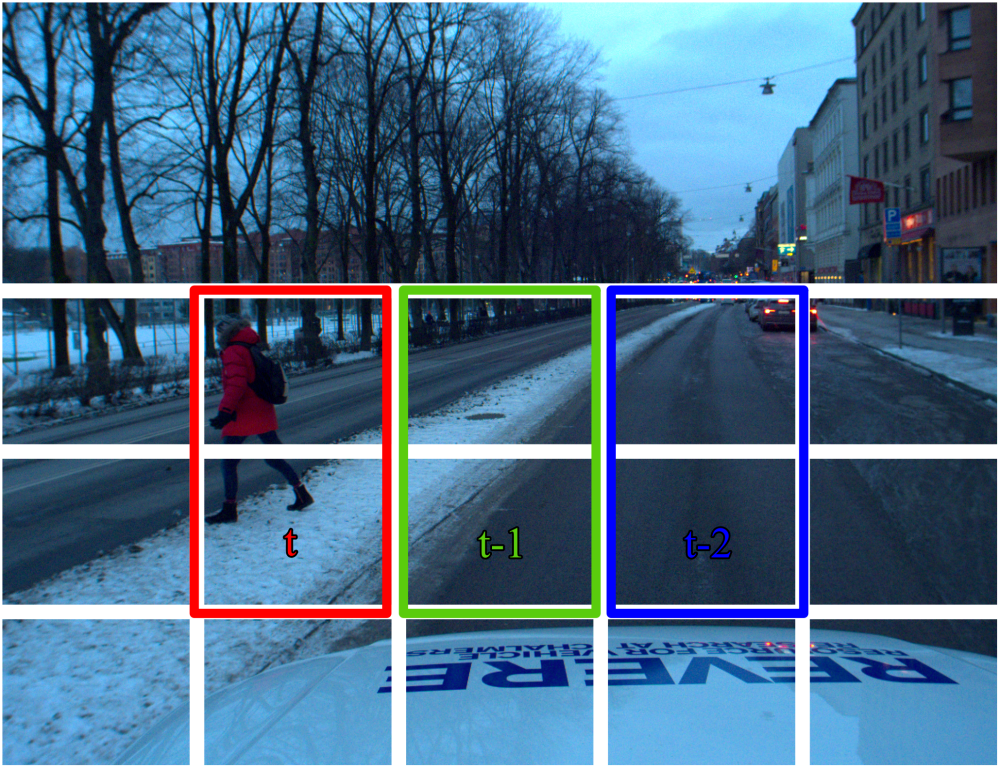}
    \caption{Representation of RoIs labelled as containing a pedestrian (rectangles) in three consecutive frames: the current time $t$, $t-1$, and $t-2$. Checking if the RoIs in frame $t$ labelled as containing a human are adjacent to the RoIs in $t-1$ and $t-2$ (DA-4) can help us detect physically impossible motions, potentially due to hallucinations in the LLM labelling.}
    \label{fig:physicalcheck}
\end{figure}

\textbf{Physical Plausibility Check (DA-4):}
The physical plausibility check was designed considering the historical data in the selected sequences. This experiment was conducted using smaller RoIs, 15 regions per frame for the selected sequences as shown in Figure~\ref{fig:physicalcheck}. This experiment is based on the physical plausibility of the detected trajectory that a pedestrian is showing through the RoIs over time. To detect false positives, we check the two previous frames to verify whether a pedestrian was also detected either in that region or in a neighboring region. This filter acts as a low-pass filter that removes high-frequency changes. The initial experiment that considered selected sequences from both datasets led to very promising results with a recall of 96.43\% and an accuracy of 85.86\%. The results and further information on the experiment, including a video that explains the experiment, can be found as supplementary materials available at this open GitHub repository~\cite{Repogithub}.   


\section{Analysis and Discussion} \label{AnalysisAndDiscussion}

The related work was investigated to find existing hallucination detection strategies and their potential applicability for an automotive context targeting ADAS/AD to support RQ1. Given the availability of resources within ADAS/AD systems, the ability to work with multimodal data, and other required factors such as the high-performance metrics made few of the suggested hallucination detection strategies inapplicable in the domain of ADAS/AD. However, the consistency checking concepts behind the hallucination detection methods such as ``selfCheckGPT~\cite{manakul2023selfcheckgpt}'', Ronanki et al.~\cite{ronanki2022chatgpt}, and ``DreamCatcher~\cite{Dreamcatcher} can be adapted as into the ADAS/AD domain by simplifying the complex prompting schemes. The proposed BO3 strategy and its adaptations that use sequential data can be considered as voting-based hallucination strategies that use consistency enhancements inspired by the literature.

The second research question (RQ2) of this study aimed at characterizing the hallucinations that can occur in an automotive context relevant for ADAS/AD, which was addressed by the results of data analysis activities DA1 and DA2. Out of the three hallucination types described in Section~\ref{sec:TypesOfHallucinations+FullImagesSingleRUn}, false negatives were considered more problematic in the context of ADAS/AD as they may potentially lead to severe consequences when not being able to detect a pedestrian. Therefore, the proceeding experiments were designed to reduce the false negatives as much as possible with higher recall values. False positives and false negatives were quite common within the selected image sequences. However, the third type, where the LLM rejected the processing of an image, hardly occurred. We also observed this to occur with the proprietary LLM that is accessed via the cloud. 

To evaluate the performance of the proposed hallucination detection mechanism BO3, DA-2 was conducted by assessing the image's RoIs three times. As shown in Table\ref{tab:da2}, the results indicated that BO3 tends to decrease the recall in GPT-4V very slightly by 0.28\%, whereas it helps in increasing recall and F1 score values of GPT-4V with both datasets. However, the interpretation of the results was quite different with LLaVA, where recall and F1 score values decreased for both datasets.

When comparing the two models with each other, it is evident that LLaVA is often inconsistent across runs, as shown in~\ref{tab:da2}. Also, as seen in Table~\ref{tab:da2_consistency}, it very seldom labels the image (yes/no) the same way across runs. 

In addition, the effect of using RoIs against the use of full images was also analyzed during DA1 and DA2 activities. The observed recall and F1 score values for both GPT-4V and LLaVA showed unpredictable deviations from each other; however, in general, the metrics showed improvements with respect to the PREPER CITY dataset. Given the fact that having RoIs significantly assisted in localizing pedestrians, which was not successful when using image captioning only, the deviations in the performance evaluation between the RoIs and full images can be neglected. 

A series of experiments focusing on sequences of data were conducted to answer the RQ-3. The adaptation of the hallucination strategies into ADAS/AD was conducted in the three different approaches THV, THV-2, and Physical Plausibility Check and analyzed as discussed under DA-3 and DA-4. Both proposed methods THV and THV-2 use temporal data to reduce hallucinations in the generated responses. With GPT-4V, the use of BO3, Single RoI, and the processing of full frames can be recommended over THV and THV-2, given the significant drops in performance as shown in Table~\ref{tab:da2}. However, with LLaVA, THV-2 has shown promising performance improvements with 17.61\% in recall and 9.89\% in F1 score compared to the BO3 strategy. 

The third approach of hallucination detection adapted to ADAS/AD context, the physical plausibility check, was evaluated with GPT-4V, showing a combined recall value of 96.43\% and 49.54\% F1 score for both Waymo and PREPER CITY datasets. This experiment was not conducted using LLaVA due to its poor performance in the previous experiments. Higher false positive rates were reported, especially with the PREPER CITY dataset, making the overall F1 score low. However, the reported higher recall values that indicate fewer false negative rates made this approach promising and suitable for future research and applications in the domain of ADAS/AD.

Threats to the validity of this study were assessed based on Feldt and Magazinius~\cite{RobertFeldt_ThreatsToValidity}. The data curation process was conducted focusing on a single pedestrian or a cohesive group of pedestrians, which reduced the number of available frames in both Waymo and PREPER CITY datasets but was intended to reduce potential unknown side-effects emerging from scattered pedestrian(s) (groups). There were different sequences from both datasets that contained driving scenarios in different seasons, including snowy conditions. However, the selected images mainly represented daytime driving scenarios, including dusk, which could have made an impact on this study. The selected images underwent a pre-processing stage by cropping and defining RoIs to support pedestrian localization, and this pre-processing may have made an impact on the overall quality of the images. The experiments that used sequential data depended on how frequently the frames were captured. For instance, if the pedestrian transitions were quick and the frequency of capturing was low, the pedestrian trajectory would not be plausible at all. The pedestrian may have sufficient time to move around the considered RoIs freely, limiting the semantical analysis of the transition. Also, while using multiple existing, widely used datasets strengthens the validity of our results, we do not have control over scenarios or sensor configurations during dataset creation. Finally, when we consider the LLMs we used, the ethical concerns, rules and regulations related to AI, technical advancements in the LLMs may limit the generalizability of these results, and hence, the integration of LLMs without any further safeguarding in a deployed LLM or in the process during its creation cannot be recommended.

\section{Conclusion and Future Work}
\label{sec:conclusion}

We assessed approaches to detect potential hallucinations when LLMs are potentially considered as part of perception tasks for ADAS/AD.
Our experiments show that the LLMs, which tend to provide negative answers, are not performing well with hallucination detection techniques that use consistency checking like Best-of-Three (BO3) as the basis. The observed performance of LLaVA under the BO3 strategy using a single video frame confirms this stochastic behavior of the model. We proposed extensions to BO3, THV and THV-2, to bridge this shortcoming of BO3. If the model tends to provide negative answers, looking back at past data in a sequence helps to improve results. However, the overall performance of the non-proprietary LLM LLaVA was significantly lower compared to the proprietary LLM. 

We also evaluated the impact of using RoIs instead of full images to support pedestrian localization. This information was very valuable in identifying plausible trajectories for pedestrians. The physical plausibility check, which adds a layer of semantics to identify plausible pedestrian movements across the RoIs in a video frame, also helps to reduce hallucinations. While this proposed approach is application-specific, it is a good fit for the domain of ADAS/AD where sequential data is easily available within the vehicles itself.

However, it was noted that both models were performing equally well in describing the images, as opposed to answering simple binary questions to locate pedestrians. This highlights the importance and relevance of our proposed extensions THV, THV-2, and physical plausibility checks. 

Our study shows the importance of using semantically relevant RoIs and sequential data for hallucination detection when considering the use of LLMs for perception tasks within the automotive context, demonstrating the potential of using state-of-the-art LLMs. We acknowledge the rapidity in the development of non-proprietary LLMs that have not only the potential to run locally without a cloud backend, but can also be fine-tuned to a particular application context. Hence, future studies need to evaluate recent adjustments to such open LLMs. Furthermore, LLMs that are made on (domain-specific) large-scale time-series data, so-called foundation models, can be combined with LLMs to better fit the application context. 

\section*{Acknowledgments}
This work has been supported by the Swedish Foundation for Strategic Research (SSF), Grant Number FUS21-0004 SAICOM and the Wallenberg AI, Autonomous Systems and Software Program (WASP) funded by the Knut and Alice Wallenberg Foundation.

\bibliographystyle{ieeetr}
\bibliography{reference}

\end{document}